\documentclass{article}
\usepackage{ijcai26}
\usepackage{times}
\usepackage[utf8]{inputenc}
\usepackage{url}
\usepackage[hidelinks]{hyperref}
\usepackage[small]{caption}
\usepackage{graphicx}
\usepackage{amsmath,amsfonts,amssymb}
\usepackage{array}
\usepackage{booktabs}
\usepackage{threeparttable}
\usepackage{multirow}
\usepackage{makecell}
\usepackage{tabularx}
\usepackage{adjustbox}
\usepackage{xcolor}
\usepackage{colortbl}
\usepackage{textcomp}
\usepackage{placeins}
\usepackage{float}
\usepackage{xspace}

\floatstyle{ruled}
\newfloat{algorithm}{tbp}{loa}
\floatname{algorithm}{Algorithm}

\hypersetup{
  breaklinks=true,
  hypertexnames=false,
  pdfstartview=Fit
}
\newcommand*{\proposed}{StepPrune\xspace}
\newcommand{\stopact}{\varnothing}
\newcommand{\softtopk}{\operatorname{SoftTopK}}
\newcommand{\sg}{\operatorname{sg}}
\newcommand{\diag}{\operatorname{diag}}

\title{StepPrune: Adaptive Sequential Visual Token Selection across Multimodal Large Language Models}

\author{
    Hansen Zhang$^1$,
    Landi He$^{1,2}$,
    Mingde Yao$^3$ {\normalfont and}
    Lijian Xu$^{1*}$
\affiliations
    $^1$Shenzhen University of Advanced Technology, Shenzhen, China\\
    $^2$School of Computer Science, Nanjing University, Nanjing, China\\
    $^3$CUHK MMLab, CPII under InnoHK, Hong Kong, China
\emails
    xulijian@suat-sz.edu.cn
}

\begin{document}
\maketitle

\begin{abstract}
\setlength{\emergencystretch}{.5em}
Visual prefixes account for a major portion of the per-layer computation in
multimodal large language models (MLLMs), making visual-token pruning a direct
approach to accelerating inference. Existing top-$K$ methods typically
evaluate tokens independently and apply a uniform budget to all inputs,
overlooking both selection-dependent interactions and variations in visual
complexity across samples. In contrast, we propose \textbf{StepPrune}, which
formulates visual-token pruning as an adaptive sequential decision process.
Conditioned on previously selected tokens and textual context, StepPrune
progressively constructs the retained subset and automatically determines its
size through a learned \textsc{Stop} action $\varnothing$. During training, a
variance-preserving noise gate provides a differentiable surrogate for the
discrete selection process, whereas during inference, unselected tokens are
physically removed before language-model prefill. A grouped selection mechanism
further extends StepPrune to high-resolution inputs. Experiments across
LLaVA-1.5, LLaVA-NeXT, Qwen2.5-VL, and InternVL3 show that StepPrune achieves
the best average normalized performance retention across all evaluated pruning
rates on LLaVA-1.5, Qwen2.5-VL, and InternVL3, while remaining competitive on
the substantially longer AnyRes prefixes of LLaVA-NeXT. On LLaVA-1.5,
StepPrune retains 94.6\% of the full-prefix normalized performance while
pruning 88.9\% of the visual tokens. At a mean retained count of 64,
StepPrune reduces prefill latency from 59.95 ms to 40.05 ms, corresponding to
a $1.50\times$ prefill speed-up.
\end{abstract}


\section{Introduction}
\label{sec:introduction}

\begin{figure*}[!t]
    \centering
    \includegraphics[width=\textwidth]{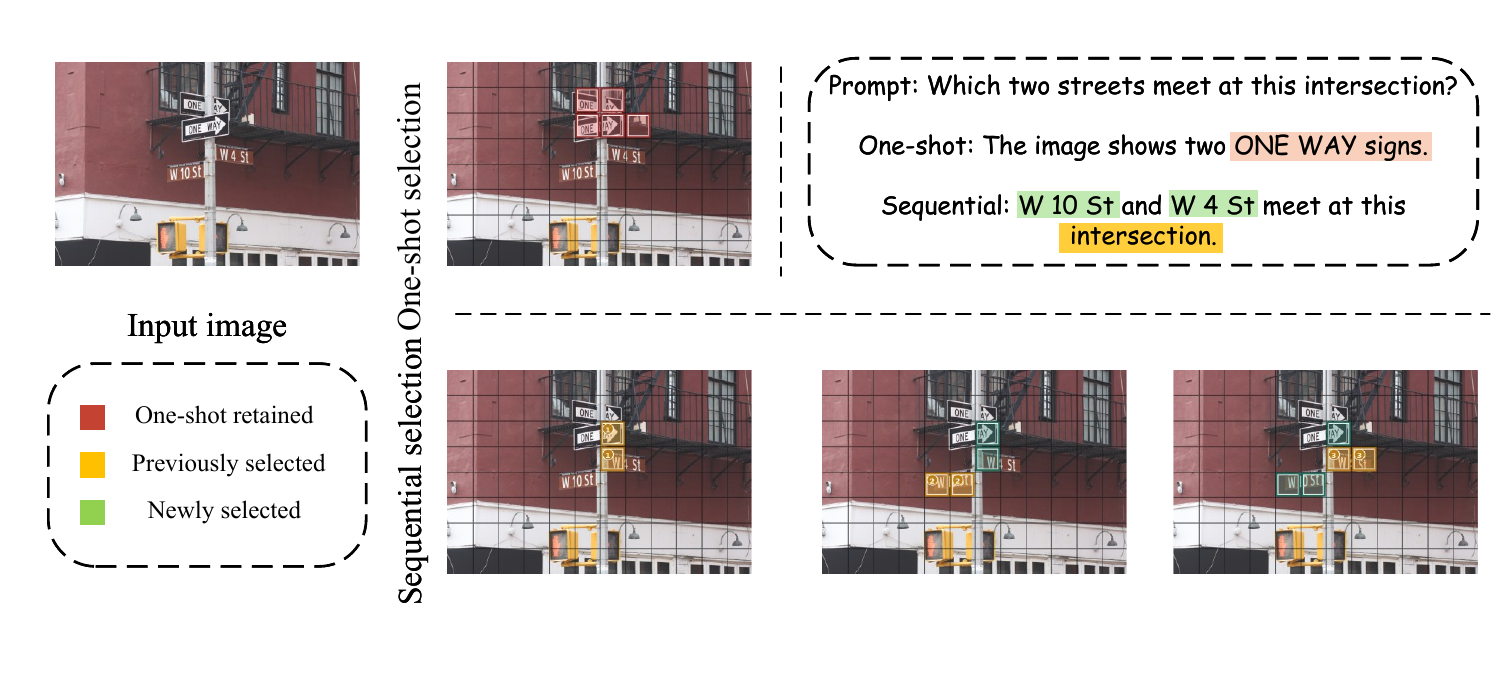}
    \caption{Comparison between one-shot and sequential visual-token selection.
    One-shot selection retains redundant sign-level tokens but misses the
    street-name evidence required by the prompt. Sequential selection
    conditions each decision on previously selected tokens and progressively
    recovers the complementary visual evidence needed to produce the correct
    answer}
    \label{fig:overview}
\end{figure*}

Multimodal large language models (MLLMs), such as LLaVA
\cite{liu2023llava}, Qwen2.5-VL \cite{bai2025qwen25vl}, and GPT-4V
\cite{openai2023gpt4v}, have become a standard architecture for
vision--language reasoning. In these models, an image is injected as a visual
prefix: a frozen vision tower \cite{radford2021clip} converts each image into
hundreds of patch tokens, which are prepended to the text prompt and consumed
by an autoregressive language model. Both inference latency and memory are
dominated by the length of this prefix rather than by the language-model
weights, because the per-layer cost of a Transformer scales superlinearly with
sequence length. Reducing the number of visual tokens that enter the language
model is therefore one of the most effective ways to improve MLLM efficiency,
and it has motivated a growing body of work on visual token compression
\cite{yao2026survey}.

Most existing approaches treat token pruning as a one-shot ranking problem: a
relevance score is computed for every visual token, for example from in-LLM
attention \cite{chen2024fastv}, vision-encoder cues
\cite{yang2025visionzip,zhang2025vispruner}, or cross-modal relevance to the
text prompt \cite{zhang2024sparsevlm}, and a top-$K$ rule retains the $K$
highest-scoring tokens. Other methods improve coverage by suppressing
duplication or maximizing conditional diversity
\cite{wen2025dart,zhang2025cdpruner}. These advances make the ranking signal
more reliable, but they do not update a candidate's marginal value after each
selection decision.
Two assumptions underlie this approach. The first is independence: each token is scored on its own, so once a token is committed to the retained set, the marginal value of the remaining candidates, especially duplicates or close neighbors of the committed one, is not re-estimated even though it has changed.
The second is a fixed compression ratio: $K$ is chosen externally
and applied uniformly, regardless of whether the input is a near-empty document
thumbnail or a dense natural scene. Real inputs vary widely in visual
complexity, and no single $K$ is appropriate for all of them. We argue that
both assumptions should be dropped, and that token selection is better framed
as a stepwise sequential decision process in which the model commits to one
token per step, conditions each step on tokens chosen earlier, and decides for
itself when to stop.

Motivated by this view, we propose \proposed, which casts selection as an
autoregressive policy
$P(S\mid X)=\prod_t\pi_\theta(a_t\mid a_{<t},X)$ realized by a pointer-style
decoder. The candidate set augments the visual tokens with a learned stop
action $\stopact$, and the episode terminates the first time $\stopact$ is
chosen, so the retained count $K$ becomes an adaptive, per-input quantity
rather than a fixed hyperparameter. Because the per-step $\arg\max$ is
non-differentiable, we pass the aggregated pointer probabilities through a
variance-preserving noise gate that mixes each visual token with isotropic
Gaussian noise under a smooth polarization of its selection score, while a
locally masked denoiser prevents cross-token leakage from bypassing the gating
signal. Figure~\ref{fig:overview} contrasts the two regimes: one-shot selection
retains redundant sign-level tokens but misses the street-name evidence
required by the prompt, whereas the stepwise policy progressively recovers
complementary cues and produces the correct answer.

Although token-wise autoregressive selection accurately models conditional
dependencies among visual tokens, its number of decoding rounds grows linearly
with visual-prefix length. For modern MLLMs with high-resolution or multi-tile
visual representations, the accumulated recurrent latency of token-wise
pointer decoding can offset the computation saved during language-model
prefill. We therefore allow the pointer decoder to select a group of visual
tokens at each round and condition subsequent decisions on the groups selected
earlier. Increasing the group size provides a controllable trade-off between
selection granularity and selector latency. This mechanism reduces the number
of decisions for long prefixes while preserving history-conditioned selection
and adaptive stopping, enabling \proposed to support the fixed-resolution
representation of LLaVA-1.5 \cite{liu2023llava}, the AnyRes representation of
LLaVA-NeXT \cite{liu2024llavanext}, the dynamic-resolution representation and
three-dimensional mRoPE of Qwen2.5-VL \cite{bai2025qwen25vl}, and the
multi-tile representation of InternVL3 \cite{zhu2025internvl3}.

We evaluate StepPrune on four MLLMs with different visual sequence organizations. StepPrune consistently preserves strong performance under different pruning rates and visual-prefix configurations. On LLaVA-1.5, it retains 94.6\% of the normalized full-prefix performance while pruning 88.9\% of the visual tokens. With an average of 64 retained tokens, it achieves a 1.50$\times$ prefill speed-up.

In summary, our main contributions are as follows:
\begin{enumerate}
    \item We formulate visual-token pruning as a history-conditioned sequential
    selection problem, where a learned STOP action jointly determines the retained
    subset and its input-adaptive size.

    \item We develop a trajectory-aware differentiable optimization scheme for
    training the discrete sequential policy from the language-modeling objective.

    \item We introduce grouped sequential selection for long visual prefixes,
    improving the scalability of StepPrune across different visual token
    organizations. Experiments on four MLLM backbones further demonstrate its applicability and efficiency.
\end{enumerate}
\section{Related Work}
\label{sec:related}

\subsection{Multimodal Large Language Models}

Modern MLLMs typically connect a visual encoder to an autoregressive
language model through an intermediate visual representation
\cite{liu2023llava,li2023blip2,alayrac2022flamingo,dai2023instructblip}.
The organization of this visual sequence varies across architectures.
LLaVA-1.5 uses a fixed-resolution patch grid, while more recent models
support AnyRes, dynamic-resolution, or multi-tile representations
\cite{liu2024llavanext,bai2025qwen25vl,zhu2025internvl3}.
These designs improve visual coverage but often produce longer visual
prefixes. As a result, the computation required during language-model
prefill increases \cite{yao2026survey}.

MLLM efficiency has also been improved through attention reduction
\cite{wu2026eas} and structural model pruning
\cite{huang2026structural}. These methods mainly reduce computation within
the model. In contrast, visual-token pruning directly shortens the visual
sequence processed by subsequent language-model layers. Our work follows
this direction and focuses on token selection across heterogeneous
visual-prefix organizations.

\begin{figure*}[!t]
    \centering
    \includegraphics[width=\textwidth]{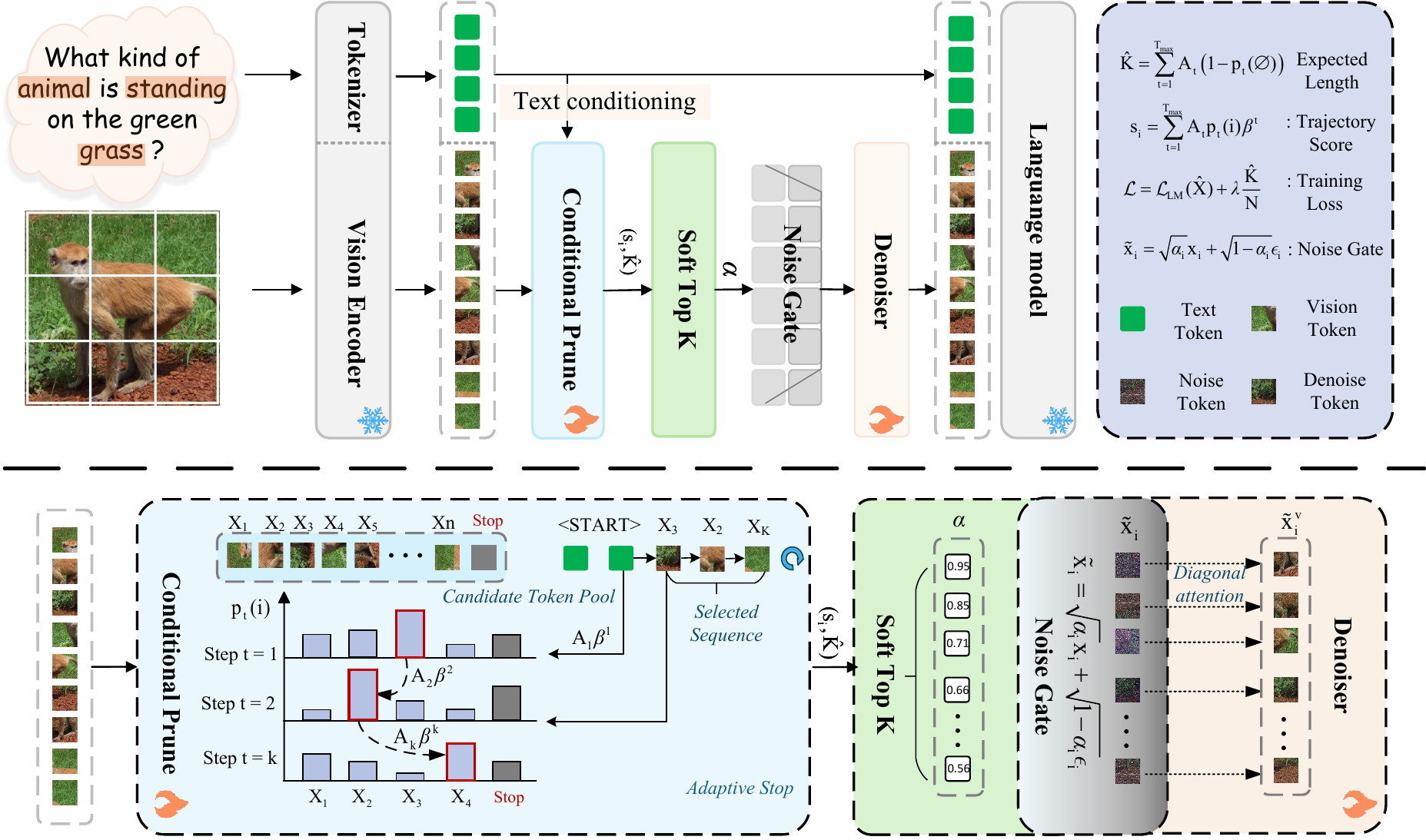}
    \caption{Overview of \proposed. Visual and text tokens are jointly encoded
    to produce cross-modal representations. The conditional pointer policy
    sequentially selects visual tokens according to the previously selected
    history and terminates with a learned STOP action. During training, the
    aggregated selection probabilities control a variance-preserving noise
    gate, while the locally masked denoiser prevents information leakage across
    visual tokens. The vision encoder and language model remain frozen.}
    \label{fig:architecture}
\end{figure*}

\subsection{Visual Token Compression and Pruning}

Token reduction was explored in vision--language Transformers before the
recent MLLM paradigm. PuMer combines text-aware pruning with token merging to
remove redundant multimodal representations \cite{cao2023pumer}. MADTP
learns instance- and layer-dependent token ratios under multimodal alignment
\cite{cao2024madtp}, and MADTP++ further combines dynamic token pruning with
hardware-aware weight pruning \cite{cao2026madtpp}. These methods introduce
adaptive computation into vision--language models, but mainly operate on
trainable Transformer architectures rather than directly reducing the visual
prefix of a frozen autoregressive MLLM.

For MLLMs, token importance is commonly estimated from attention or
cross-modal relevance. FastV uses attention statistics from the language model
\cite{chen2024fastv}, while SparseVLM exploits visual--text relevance for
progressive pruning \cite{zhang2024sparsevlm}. FitPrune derives layer-wise
pruning ratios by matching attention distributions
\cite{ye2025fitprune}. LLaVA-PruMerge and VisPruner instead exploit
vision-encoder attention together with feature similarity or spatial
redundancy \cite{shang2024prumerge,zhang2025vispruner}. More recent methods
further improve token scoring through holistic context, head-wise attention,
or approximation error
\cite{zou2025holov,zhu2026hawk,ma2026apet}. Despite different scoring cues,
these methods do not explicitly update the value of remaining candidates
according to the subset already selected.

Beyond token saliency, several methods focus on preserving coverage and
reducing redundancy. VisionZip retains dominant visual tokens and merges
contextual information \cite{yang2025visionzip}. DART suppresses redundant
tokens, while CDPruner further incorporates instruction-conditioned diversity
into subset selection \cite{wen2025dart,zhang2025cdpruner}. VFlowOpt combines
attention relevance, patch entropy, and information-flow cues
\cite{yang2025vflowopt}. PyramidDrop adopts a different strategy and
progressively reduces the visual sequence across language-model layers
\cite{xing2025pyramiddrop}.

\subsection{Learned and Adaptive Token Selection}

A more recent line of work learns the pruning policy from data rather than
relying on predefined scoring rules. In vision Transformers, DynamicViT learns
instance-dependent token sparsification \cite{rao2021dynamicvit}, A-ViT
introduces token-wise halting \cite{yin2022avit}, and ATS performs
input-adaptive token sampling \cite{fayyaz2022ats}. These methods show that
the amount of computation can be adjusted according to the input, but their
decisions are made within the vision encoder.

Adaptive token selection has also been explored in MLLMs. Glimpse Prune learns
a token-importance predictor with localization supervision
\cite{zeng2026glimpse}. ATP-LLaVA predicts per-instance retention ratios
between language-model layers \cite{ye2025atpllava}, while p-MoD formulates
token processing as Mixture-of-Depths routing \cite{zhang2025pmod}.
TopV and AdaLLaVA further adapt token processing according to the input or
runtime budget \cite{yang2025topv,xu2025adallava}. These methods relax the
use of a uniform token budget. However, they do not explicitly model the
retained subset as a sequence of decisions conditioned on earlier selections.

\proposed differs along this conditioning dimension. It constructs the
retained subset sequentially and recomputes the selection distribution after
each decision. A learned STOP action determines when the process terminates,
so subset membership and retained length are produced by the same policy.
We use the noise-gated surrogate of AutoSelect \cite{he2026autoselect} to
optimize the discrete selection process. For long visual prefixes, the same
policy is executed in groups to reduce the number of sequential decisions.

\section{Method}
\label{sec:method}

We formulate visual-token pruning as a conditional sequential selection
problem. Instead of assigning all visual tokens independent scores and then
applying a fixed top-$K$ rule, \proposed constructs the retained subset
progressively. Each selection depends on the tokens chosen earlier, while a
learned STOP action determines when the selection process terminates. As shown
in Fig.~\ref{fig:architecture}, the policy is implemented with a cross-modal
pointer decoder. A trajectory-based noise-gated pathway provides the
differentiable signal required to train the discrete selection process. For
high-resolution inputs with long visual prefixes, we reduce recurrent
selection cost by retaining five visual tokens at each pointer step.

\subsection{Problem Formulation}
\label{sec:preliminaries}

\paragraph{Setup and Task.}
Given an image and its instruction $Q$, the vision encoder produces a sequence
of visual tokens $X=[x_1,\ldots,x_N]\in\mathbb{R}^{N\times D}$, where $N$
denotes the number of visual tokens and $D$ is the feature dimension. The
visual features are subsequently mapped to the language-model space by the
standard multimodal projector and combined with the text prompt.

Conventional visual-token pruning methods commonly assign each token an
importance score and retain the highest-scoring tokens under a prescribed
budget:
\begin{equation}
s_i=f_\phi(x_i,Q),\qquad
\mathcal{S}
=
\operatorname{top\text{-}K}
\left(
\{s_i\}_{i=1}^{N},K_0
\right),
\label{eq:independent}
\end{equation}
where $K_0$ is specified before selection. Once the scores are computed, the
remaining candidates are not explicitly re-evaluated according to the tokens
that have already been retained.

We instead represent pruning by an ordered action trajectory
$A=(a_1,\ldots,a_T)$, where
$a_t\in\{1,\ldots,N\}\cup\{\stopact\}$. Its probability is factorized as
\begin{equation}
P_\theta(A\mid X,Q)
=
\prod_{t=1}^{T}
\pi_\theta
\left(
a_t\mid a_{<t},X,Q
\right),
\label{eq:policy}
\end{equation}
where $a_{<t}$ denotes the preceding selection history. The action
$\stopact$ terminates the trajectory. If the first STOP action occurs at
step $T^\star$, the retained sequence is
$\mathcal{S}(A)=(a_1,\ldots,a_{T^\star-1})$ and its length is
$K=T^\star-1$. Both the retained tokens and their number are therefore
determined by the same conditional policy. We denote the pointer distribution
over the $N+1$ candidates at step $t$ by $p_t$.

\subsection{Conditional Sequential Pointer Policy}
\label{sec:pointer}

The policy in Eq.~\eqref{eq:policy} is implemented with a cross-modal input
encoder and a pointer decoder. The encoder introduces instruction information
into the visual representation, while the decoder constructs the retained
subset over multiple selection steps.

\paragraph{Cross-Modal Input Encoder.}
The input encoder contains two pre-norm Vision Transformer blocks. The text
embeddings are first projected into the visual feature space as
$\widetilde{E}_{\mathrm{text}}
=g_{\mathrm{text}}\operatorname{LN}(W_pE_{\mathrm{text}})$, where
$E_{\mathrm{text}}\in
\mathbb{R}^{L_{\mathrm{text}}\times D_{\mathrm{text}}}$
contains the prompt embeddings and
$W_p\in\mathbb{R}^{D\times D_{\mathrm{text}}}$
is a learned projection. The projected text embeddings and visual tokens are
then jointly encoded:
\begin{equation}
H
=
\operatorname{Enc}
\left(
[X;\widetilde{E}_{\mathrm{text}}]
\right)_{[1:N]}
\in\mathbb{R}^{N\times D}.
\label{eq:encoder}
\end{equation}
The coefficient $g_{\mathrm{text}}\in[0,1]$ is gradually increased during
early training. After joint encoding, only the visual positions are retained
for selection. The text embeddings influence $H$ through cross-modal
self-attention but are not themselves selectable.

\paragraph{Learned STOP Action.}
A learned embedding
$s_{\mathrm{stop}}\in\mathbb{R}^{D}$ is appended to the encoded visual tokens,
forming the candidate memory
$M=[H;s_{\mathrm{stop}}]\in\mathbb{R}^{(N+1)\times D}$. The STOP action
therefore occupies the same pointer space as the visual candidates. The policy
can terminate according to the current input and selection history rather than
a predefined per-image token count.

\paragraph{Pointer Decoder.}
We adopt pointer attention \cite{vinyals2015pointer} as the selection
primitive. Each decoder layer contains causal self-attention over the
selection history, cross-attention to the candidate memory $M$, and a
feed-forward network. The cross-attention logits are reused as pointer logits
over the $N+1$ candidates.

The decoder is initialized with a learned START embedding
$q_1=s_{\mathrm{start}}$. Previously selected visual positions are masked so
that each token can be selected at most once. At step $t$, the pointer update
is
\begin{equation}
\begin{aligned}
\ell_t
&=
\operatorname{PtrAttn}(q_t,M)
+\mathrm{Mask}_t,\\
p_t
&=
\operatorname{softmax}(\ell_t/\tau),\\
a_t
&=
\arg\max_{j\in\{1,\ldots,N+1\}}
\ell_t(j),\\
\mathrm{Mask}_{t+1}(i)
&=
\begin{cases}
-\infty, & i=a_t\leq N,\\
\mathrm{Mask}_{t}(i), & \text{otherwise},
\end{cases}\\
\widetilde p_t
&=
e_{a_t}-\sg(p_t)+p_t,\\
q_{t+1}
&=
M^\top\widetilde p_t .
\end{aligned}
\label{eq:pointer-step}
\end{equation}
Here, $\tau$ denotes the pointer temperature,
$e_{a_t}$ is the one-hot representation of the selected action, and
$\sg(\cdot)$ denotes stop-gradient. The forward path uses the hard action
$a_t$, whereas $\widetilde p_t$ provides a straight-through path through the
pointer distribution.

The selected embedding is incorporated into the causal decoder history before
the next step. The pointer distribution is therefore recomputed after each
selection. As the retained subset changes, the remaining candidates are
evaluated under the updated history. Selection terminates when
$a_t=\stopact$ or when the maximum number of selection steps is reached.

\subsection{Trajectory-Aware Differentiable Optimization}
\label{sec:noise-gate}

The hard decision in Eq.~\eqref{eq:pointer-step} cannot be optimized directly
with the language-modeling objective. We therefore aggregate the pointer
distributions along the selection trajectory and obtain a continuous
per-token training signal.

\paragraph{Trajectory Aggregation.}
Let $A_t$ denote the probability that the selection process remains active
before step $t$:
\begin{equation}
A_t
=
\prod_{u=1}^{t-1}
\left(
1-p_u(\stopact)
\right),
\qquad
A_1=1.
\label{eq:alive}
\end{equation}
For each visual token $i$, its pointer probabilities are accumulated across
the trajectory:
\begin{equation}
s_i
=
\sum_{t=1}^{T_{\max}}
A_t\,p_t(i)\,\beta^t,
\qquad
\beta\in(0,1),
\label{eq:aggregate}
\end{equation}
where $\beta$ is a geometric discount factor. The term $A_t$ reduces the
contribution of later steps when the policy is increasingly likely to have
terminated. The factor $\beta^t$ places greater weight on tokens favored
earlier in the trajectory. The resulting $s_i$ summarizes the selection
preference accumulated over the complete conditional process.

The corresponding differentiable retained count is
\begin{equation}
\overline K
=
\sum_{t=1}^{T_{\max}}
A_t
\left(
1-p_t(\stopact)
\right).
\label{eq:soft-count}
\end{equation}
This quantity is used during optimization. At inference, the actual retained
length is determined by the first hard STOP action.

\paragraph{Score Polarization.}
The trajectory scores in Eq.~\eqref{eq:aggregate} are continuous and do not
directly form a binary retention mask. We map them to $[0,1]$ with a
differentiable top-$k$ operator \cite{xie2020softtopk}:
\begin{equation}
\begin{aligned}
\alpha_i
&=
\softtopk_{k,\tau}(s)_i
\in[0,1],\\
k
&=
\sg
\left(
\frac{1}{B}
\sum_{b=1}^{B}
\overline K_b
\right),
\end{aligned}
\label{eq:polarization}
\end{equation}
where $B$ denotes the batch size. The detached value of $k$ calibrates the
polarized scores to the current expected retained length. It is used only for
the continuous training pathway and does not prescribe the inference-time
token count of an individual input.

\paragraph{Variance-Preserving Noise Gate.}
The polarized score $\alpha_i$ is used to construct a continuous proxy for
token retention through variance-preserving noise gating
\cite{he2026autoselect}:
\begin{equation}
\widetilde{x}_i
=
\sqrt{\alpha_i}\,x_i
+
\sqrt{1-\alpha_i}\,\epsilon_i,
\qquad
\epsilon_i\sim\mathcal{N}(0,I_D).
\label{eq:vp-gate}
\end{equation}
When $\alpha_i$ approaches one, the original visual token is largely
preserved. When $\alpha_i$ approaches zero, it is progressively replaced by
Gaussian noise. Since $\alpha_i$ is obtained from the trajectory score in
Eq.~\eqref{eq:aggregate}, the gating strength reflects the complete
history-conditioned selection process.

\paragraph{Locally Masked Denoiser.}
Noise injection perturbs the visual feature distribution. We therefore apply
one Vision Transformer block after the noise gate,
$\widehat X=\operatorname{Denoise}(\widetilde X;\mathrm{mask}=\diag)$.
Self-attention in this block is restricted to the diagonal. Each token can be
refined independently, but information cannot be transferred between visual
positions. A strongly noised token therefore cannot recover its content from
neighboring tokens.

\paragraph{Train--Inference Alignment.}
During training, the language model receives the full-length gated sequence
$\widehat X$ in its original visual order. During inference, the noise gate
and denoiser are removed. The pointer policy produces a hard retained subset,
and the corresponding original visual tokens are physically gathered before
language-model prefill.

As the polarized scores become sharper, $\alpha_i$ approaches a
keep-or-suppress indicator induced by the selection trajectory. The
noise-gated pathway therefore provides a continuous approximation to the hard
pruning operation. The selected indices are restored to their native visual
order before projection so that the positional convention of the underlying
MLLM is preserved.

\subsection{Training Objective}
\label{sec:objective}

The gated visual sequence is passed through the frozen multimodal projector
and concatenated with the prompt embeddings. We optimize the
language-modeling objective together with a light length regularizer:
\begin{equation}
\mathcal{L}
=
\mathcal{L}_{\mathrm{LM}}(\widehat X)
+
\lambda
\frac{\overline K}{N},
\qquad
\lambda=0.03.
\label{eq:objective}
\end{equation}
The language-modeling term encourages the selector to preserve the visual
evidence required by the task. The length regularizer discourages
unnecessarily long selection trajectories through the STOP probabilities in
Eq.~\eqref{eq:soft-count}. The vision tower, multimodal projector, and
language model remain frozen. Only the cross-modal encoder, pointer decoder,
locally masked denoiser, text projection, and learned START and STOP
embeddings are optimized.

\subsection{Grouped Execution for High-Resolution Inputs}
\label{sec:long-prefix}

The standard policy updates the pointer distribution after every retained
visual token. This provides fine-grained history conditioning, but the number
of recurrent pointer updates grows with the retained length. High-resolution
inputs can produce substantially longer visual prefixes, making token-wise
execution unnecessarily expensive.

For these inputs, each non-STOP pointer step retains five visual tokens. At
round $r$, the pointer first evaluates all unselected visual candidates and
the STOP action. If STOP has the largest logit, the process terminates.
Otherwise, the five highest-scoring unselected visual tokens are retained:
\begin{equation}
U_r
=
\operatorname{TopG}
\left(
\left\{
\ell_r(i):
i\notin\mathcal{S}_{r-1},\,
i\leq N
\right\},
5
\right).
\label{eq:grouped-execution}
\end{equation}
The selected group is incorporated into the causal selection history before
the next pointer update. The pointer distribution is then recomputed using
the updated retained evidence. The standard token-wise setting corresponds to
one retained token per step, whereas the high-resolution setting retains five.
For a final retained length $K$, this reduces the number of recurrent pointer
updates from approximately $K+1$ to $\lceil K/5\rceil+1$ while preserving the
same STOP decision and history-conditioned selection rule.

Algorithm~\ref{alg:stepprune} summarizes the inference-time selection
procedure. The group size is set to $g=1$ for the standard token-wise setting
and $g=5$ for high-resolution inputs.

\begin{algorithm}[!ht]
\small
\caption{Inference-time adaptive sequential selection in \proposed}
\label{alg:stepprune}
\textbf{Require:} Visual tokens $X$, text embeddings $E_{\mathrm{text}}$,
group size $g$, cap $T_{\max}$, temperature $\tau$\par
\textbf{Ensure:} Retained visual prefix $X_S$
\begin{enumerate}
\item $\widetilde E_{\mathrm{text}}\gets
g_{\mathrm{text}}\operatorname{LN}(W_pE_{\mathrm{text}})$;
$H\gets\operatorname{Enc}([X;\widetilde E_{\mathrm{text}}])_{1:N}$.
\item $M\gets[H;s_{\mathrm{stop}}]$; $S\gets\emptyset$;
$q_1\gets s_{\mathrm{start}}$; $\mathrm{Mask}_1\gets0$.
\item \textbf{For} $r=1,\ldots,T_{\max}$:
  \begin{enumerate}
  \item $\ell_r\gets\operatorname{PtrAttn}(q_r,M)+\mathrm{Mask}_r$.
  \item \textbf{If} $\arg\max_j\ell_r(j)=N+1$, \textbf{break} (STOP).
  \item $U_r\gets
  \operatorname{TopG}(\{\ell_r(i):i\notin S,\ i\le N\},g)$.
  \item $S\gets S\cup U_r$; mask all indices in $U_r$.
  \item Append $M[U_r]$ to the causal pointer history and update $q_{r+1}$.
  \end{enumerate}
\item Restore $S$ to the native visual order.
\item $X_S\gets\operatorname{Gather}(X,S)$; \textbf{return} $X_S$.
\end{enumerate}
\end{algorithm}

\section{Experiments}
\label{sec:experiments}

We perform experiments to evaluate whether \proposed preserves task-relevant
visual information, generalizes across different visual-token organizations,
adapts the retained budget to individual inputs, and yields practical
inference gains under aggressive token pruning. We first evaluate four
single-image MLLMs covering fixed-resolution, dynamic-resolution, AnyRes, and
multi-tile visual representations. We then analyze adaptive budget allocation
and sequential selection dynamics. Finally, we examine computational
efficiency, component effectiveness, and hyperparameter sensitivity.

\subsection{Experimental Setup}
\label{sec:setup}

\subsubsection{Models and Evaluation Scope}

We evaluate \proposed on four representative MLLMs with substantially
different visual-token organizations. LLaVA-1.5-7B
\cite{liu2023llava} uses CLIP ViT-L/14 \cite{radford2021clip} and produces a
fixed visual prefix of $N=576$ tokens. Qwen2.5-VL-7B
\cite{bai2025qwen25vl} adopts dynamic-resolution visual encoding together
with three-dimensional mRoPE. LLaVA-NeXT-7B
\cite{liu2024llavanext} represents the high-resolution AnyRes setting and
produces $N=2880$ visual tokens in the evaluated configuration. InternVL3-8B
\cite{zhu2025internvl3} further tests multi-tile high-resolution
representations. Together, these backbones allow us to evaluate whether the
conditional selection policy transfers beyond a single visual encoder or
tokenization scheme.

\subsubsection{Benchmarks}

We evaluate on eight standard multimodal benchmarks. General visual question
answering and reasoning are measured with GQA \cite{hudson2019gqa}, VQAv2
\cite{goyal2017vqa}, and ScienceQA-IMG \cite{lu2022scienceqa}. MMBench and
MMBench-CN \cite{liu2024mmbench} evaluate general and multilingual
multimodal understanding, while MME \cite{fu2023mme} provides a comprehensive
perception and cognition benchmark. TextVQA \cite{singh2019textvqa} measures
fine-grained text understanding, and POPE \cite{li2023pope} evaluates object
hallucination. Because baseline availability differs across backbones, each
main-result table reports the largest common benchmark subset available for
that backbone. We follow the official evaluation protocols and use directly
comparable baseline results whenever available.

\subsubsection{Baselines and Implementation Details}

We compare \proposed with representative visual-token reduction methods,
including FastV \cite{chen2024fastv}, SparseVLM
\cite{zhang2024sparsevlm}, PyramidDrop (PDrop)
\cite{xing2025pyramiddrop}, VisionZip \cite{yang2025visionzip}, ApET
\cite{ma2026apet}, HoloV \cite{zou2025holov}, DivPrune
\cite{alvar2025divprune}, DART \cite{wen2025dart}, and ERA
\cite{wang2026era}. These baselines cover attention-based ranking,
text-conditioned sparsification, progressive reduction, token merging,
diversity-aware selection, and approximation-error-guided compression.
Comparisons are performed within the same backbone and at matched pruning
rates.

We train the selector encoder, pointer decoder, local denoiser, text
projection, and learned START and STOP embeddings while freezing the vision
tower, multimodal projector, and language model. For LLaVA-1.5-7B, we use
$T_{\max}=288$, $\beta=0.9$, and $\lambda=0.03$ for the default
configuration. The reported operating points are obtained with separately
trained selectors under different length pressures; therefore, the displayed
retained counts are test-split means rather than fixed per-image budgets. We
use AdamW with a cosine learning-rate schedule from $1\times10^{-5}$ to
$1\times10^{-6}$ and an effective batch size of 16. All models are trained on
a single NVIDIA RTX PRO 6000 Blackwell GPU. For Qwen2.5-VL, LLaVA-NeXT, and
InternVL3, we retain the same frozen-backbone protocol and use grouped
sequential selection with $g=5$ for the longer visual prefixes. Pruning rates
are computed relative to the native visual-prefix length of each backbone.

\subsection{Main Results}
\label{sec:main-results}

Single-image evaluation is the primary test of whether the sequential policy
preserves the visual evidence required by downstream tasks under different
token budgets. We begin with LLaVA-1.5-7B and then extend the evaluation to
Qwen2.5-VL-7B, LLaVA-NeXT-7B, and InternVL3-8B. Within each table, methods
are compared under the same backbone and matched pruning rates. The best and
second-best pruned results are shown in bold and underlined, respectively.

\subsubsection{Results on LLaVA-1.5-7B}
\label{sec:llava-results}

Table~\ref{tab:llava-main} first examines how \proposed behaves as the visual
budget becomes progressively tighter. The evaluated operating points retain
mean token counts of 192, 128, and 64, corresponding to pruning rates of
66.7\%, 77.8\%, and 88.9\%. While the baselines use fixed per-image budgets,
the counts reported for \proposed are test-split means produced by its adaptive
STOP policy.

The benefit of \proposed becomes more visible as the retained budget
decreases. Its average performance retention changes from 97.9\% at 66.7\%
pruning to 96.7\% and 94.6\% at 77.8\% and 88.9\%, while the margin over
VisionZip increases from 0.1 to 0.3 and 2.6 percentage points. POPE provides
the clearest example of this trend: the gain over VisionZip grows from
0.4 points at 66.7\% pruning to 2.0 points at 77.8\% and 6.1 points at
88.9\%. This suggests that history-conditioned selection becomes increasingly
useful when a compact visual prefix must preserve complementary evidence.
The remaining benchmarks exhibit a mixture of gains and losses, indicating
an overall advantage rather than uniform dominance across tasks.

\begin{table*}[!t]
\centering
\caption{Results on LLaVA-1.5-7B. ``Avg.'' is the mean ratio of each
benchmark score to the full-prefix upper bound. For \proposed, $K$ denotes
the test-split mean selected count. Best and second-best pruned results at
each pruning rate are shown in bold and underlined, respectively.}
\label{tab:llava-main}

\begin{adjustbox}{max width=\linewidth}
\begin{tabular}{lccccccccc}
\toprule
Method &
GQA &
MMB &
MMBCN &
MME &
POPE &
SQA &
VQAv2 &
TextVQA &
Avg. (\%) \\
\midrule

\rowcolor{gray!10}
\multicolumn{10}{c}{\textsc{Upper Bound: 576 Tokens (100\%)}} \\

Vanilla
& 61.9
& 64.7
& 58.1
& 1862
& 85.9
& 69.5
& 78.5
& 58.2
& 100.0 \\

\midrule

\rowcolor{gray!10}
\multicolumn{10}{c}{\textsc{Token Pruning Rate = 66.7\%}} \\

FastV
& 52.7
& 61.2
& 53.5
& 1612
& 64.8
& 67.3
& 67.1
& 52.5
& 88.3 \\

SparseVLM
& 57.6
& 62.5
& \textbf{58.6}
& 1721
& 83.6
& 69.1
& 75.6
& 56.1
& 96.5 \\

PDrop
& 57.3
& \textbf{63.6}
& 56.8
& \underline{1797}
& 82.3
& \underline{69.2}
& 75.1
& \underline{56.5}
& 96.7 \\

VisionZip
& 59.3
& 63.0
& 57.3
& 1783
& 85.3
& 68.9
& \textbf{76.8}
& \textbf{57.3}
& \underline{97.8} \\

ApET
& \textbf{60.2}
& 63.4
& \underline{57.9}
& \textbf{1808}
& \textbf{86.3}
& 68.5
& 76.2
& 54.4
& 97.7 \\

\rowcolor{orange!8}
\proposed
& \underline{59.5}
& \underline{63.5}
& 57.4
& 1773
& \underline{85.7}
& \textbf{69.9}
& \underline{76.6}
& \underline{56.5}
& \textbf{97.9} \\

\midrule

\rowcolor{gray!10}
\multicolumn{10}{c}{\textsc{Token Pruning Rate = 77.8\%}} \\

FastV
& 49.6
& 56.1
& 55.9
& 1490
& 59.6
& 60.2
& 61.8
& 50.6
& 83.1 \\

SparseVLM
& 56.0
& 60.0
& 51.1
& 1696
& 80.5
& 67.1
& 73.8
& 54.9
& 92.6 \\

PDrop
& 57.1
& 61.6
& 56.6
& 1761
& 82.3
& 68.4
& 72.9
& \underline{56.6}
& 95.5 \\

VisionZip
& \underline{57.6}
& 62.0
& \underline{56.7}
& \underline{1762}
& 83.2
& \underline{68.9}
& \underline{75.6}
& \textbf{56.8}
& 96.4 \\

ApET
& \textbf{58.9}
& \underline{62.3}
& 56.4
& \textbf{1801}
& \textbf{86.1}
& 68.7
& 75.1
& 53.9
& \underline{96.6} \\

\rowcolor{orange!8}
\proposed
& 57.3
& \textbf{63.2}
& \textbf{57.1}
& 1749
& \underline{85.2}
& \textbf{69.8}
& \textbf{75.7}
& 55.3
& \textbf{96.7} \\

\midrule

\rowcolor{gray!10}
\multicolumn{10}{c}{\textsc{Token Pruning Rate = 88.9\%}} \\

FastV
& 46.1
& 48.0
& 52.7
& 1256
& 48.0
& 51.1
& 55.0
& 47.8
& 73.6 \\

SparseVLM
& 52.7
& 56.2
& 46.1
& 1505
& 75.1
& 62.2
& 68.2
& 51.8
& 85.6 \\

PDrop
& 47.5
& 58.8
& 50.5
& 1561
& 55.9
& \underline{69.0}
& 69.2
& 50.6
& 84.7 \\

VisionZip
& 55.1
& 60.1
& 50.4
& 1690
& 77.0
& \underline{69.0}
& 72.4
& \textbf{55.5}
& 92.0 \\

ApET
& \textbf{56.9}
& \underline{61.2}
& \underline{54.4}
& \textbf{1714}
& \textbf{84.4}
& 68.9
& \underline{72.5}
& 53.0
& \underline{94.1} \\

\rowcolor{orange!8}
\proposed
& \underline{56.4}
& \textbf{61.8}
& \textbf{55.4}
& \underline{1698}
& \underline{83.1}
& \textbf{69.3}
& \textbf{72.9}
& \underline{54.7}
& \textbf{94.6} \\

\bottomrule
\end{tabular}
\end{adjustbox}
\end{table*}

\subsubsection{Results on Qwen2.5-VL-7B}
\label{sec:qwen-results}

We  test whether the learned policy transfers to the dynamic-resolution
visual representation of Qwen2.5-VL-7B. Unlike LLaVA-1.5, the number and
organization of visual tokens vary with the input, providing a substantially
different setting for sequential selection.

As shown in Table~\ref{tab:qwen-main}, \proposed retains 97.8\%, 95.0\%,
and 90.9\% of the full-prefix performance at the three pruning rates,
exceeding HoloV by 3.5, 4.2, and 3.9 percentage points, respectively.
Importantly, the advantage remains substantial as pruning becomes more
aggressive rather than disappearing after transfer to the new backbone.
At 88.9\% pruning, for example, TextVQA increases from 61.8 with HoloV to
69.3 with \proposed, while gains are also maintained on MMBench, MME, POPE,
and ScienceQA-IMG. These results indicate that the conditional selection
policy is not tied to the fixed visual grid or the LLaVA--CLIP
representation.

\begin{table*}[!t]
\centering
\caption{Results on Qwen2.5-VL-7B. The definition of ``Avg.'' follows
Table~\ref{tab:llava-main}. Best and second-best pruned results at each
pruning rate are shown in bold and underlined, respectively.}
\label{tab:qwen-main}

\begin{adjustbox}{max width=\linewidth}
\begin{tabular}{lcccccc}
\toprule
Method &
MMB &
MME &
POPE &
SQA &
TextVQA &
Avg. (\%) \\
\midrule

\rowcolor{gray!10}
\multicolumn{7}{c}{\textsc{Upper Bound: Full Visual Tokens (100\%)}} \\

Vanilla
& 82.8
& 2304
& 86.1
& 84.7
& 84.8
& 100.0 \\

\midrule

\rowcolor{gray!10}
\multicolumn{7}{c}{\textsc{Token Pruning Rate = 66.7\%}} \\

FastV
& 75.7
& 2072
& 82.2
& 78.5
& \underline{77.9}
& 92.3 \\

HoloV
& \underline{78.3}
& \underline{2093}
& \underline{85.0}
& \underline{79.8}
& \textbf{78.9}
& \underline{94.3} \\

\rowcolor{orange!8}
\proposed
& \textbf{79.5}
& \textbf{2300}
& \textbf{85.3}
& \textbf{86.9}
& 77.5
& \textbf{97.8} \\

\midrule

\rowcolor{gray!10}
\multicolumn{7}{c}{\textsc{Token Pruning Rate = 77.8\%}} \\

FastV
& 74.9
& 2036
& 80.7
& 78.0
& 69.0
& 89.2 \\

HoloV
& \underline{76.5}
& \underline{2043}
& \underline{82.3}
& \underline{79.8}
& \underline{70.3}
& \underline{90.8} \\

\rowcolor{orange!8}
\proposed
& \textbf{78.8}
& \textbf{2222}
& \textbf{83.0}
& \textbf{83.7}
& \textbf{74.8}
& \textbf{95.0} \\

\midrule

\rowcolor{gray!10}
\multicolumn{7}{c}{\textsc{Token Pruning Rate = 88.9\%}} \\

FastV
& 69.2
& 1940
& 78.6
& 77.4
& 60.3
& 84.3 \\

HoloV
& \underline{72.4}
& \underline{2006}
& \underline{80.7}
& \underline{79.5}
& \underline{61.8}
& \underline{87.0} \\

\rowcolor{orange!8}
\proposed
& \textbf{74.8}
& \textbf{2122}
& \textbf{80.8}
& \textbf{81.6}
& \textbf{69.3}
& \textbf{90.9} \\

\bottomrule
\end{tabular}
\end{adjustbox}
\end{table*}

\subsubsection{Results on LLaVA-NeXT-7B}
\label{sec:llavanext-results}

We  evaluate \proposed on the substantially longer visual prefix of
LLaVA-NeXT-7B. In the evaluated AnyRes configuration, the visual sequence
increases from 576 tokens in LLaVA-1.5 to 2880 tokens, allowing us to examine
whether sequential selection remains effective when the candidate space
becomes much larger.

At 77.8\% pruning, \proposed retains 97.1\% of the full-prefix performance,
only 0.5 percentage points below ApET, while achieving the strongest results
on GQA, MMBench, MME, and VQAv2. Performance remains competitive as the
compression ratio increases, reaching 92.1\% and 88.2\% average retention at
88.9\% and 94.4\% pruning, respectively. The degradation, however, is not
uniform across tasks. At the two tightest budgets, TextVQA drops to 49.6 and
43.3, whereas GQA, MME, and POPE remain comparatively robust. This contrast
suggests that the policy can scale to long AnyRes prefixes, but preserving
small and spatially localized textual evidence becomes increasingly difficult
under extreme compression.

\begin{table*}[!t]
\centering
\caption{Results on LLaVA-NeXT-7B. ``Avg.'' is the mean ratio of each
benchmark score to the full-prefix upper bound over the available common
benchmarks. Best and second-best pruned results at each pruning rate are
shown in bold and underlined, respectively.}
\label{tab:llavanext-main}

\begin{adjustbox}{max width=\linewidth}
\begin{tabular}{lcccccccc}
\toprule
Method &
GQA &
MMB &
MMBCN &
MME &
POPE &
VQAv2 &
TextVQA &
Avg. (\%) \\
\midrule

\rowcolor{gray!10}
\multicolumn{9}{c}{\textsc{Upper Bound: 2,880 Tokens (100\%)}} \\

Vanilla
& 64.2
& 67.4
& 60.6
& 1851
& 86.5
& 81.8
& 61.3
& 100.0 \\

\midrule

\rowcolor{gray!10}
\multicolumn{9}{c}{\textsc{Token Pruning Rate = 77.8\%}} \\

SparseVLM
& 60.3
& \underline{65.8}
& 58.5
& 1773
& 84.2
& 77.1
& 57.8
& 95.7 \\

PDrop
& 60.6
& 65.5
& 58.5
& 1781
& 83.7
& 78.3
& 57.4
& 95.8 \\

VisionZip
& \underline{63.0}
& 65.0
& 52.2
& 1714.3
& \underline{86.5}
& 75.9
& \textbf{61.9}
& 95.3 \\

ApET
& \underline{63.0}
& 65.3
& \textbf{59.3}
& \underline{1815}
& \textbf{87.2}
& \underline{79.2}
& 57.9
& \underline{97.6} \\

\rowcolor{orange!8}
\proposed
& \textbf{64.0}
& \textbf{67.3}
& \underline{59.2}
& \textbf{1826}
& \underline{86.5}
& \textbf{79.8}
& \underline{59.6}
& \textbf{98.7} \\

\midrule

\rowcolor{gray!10}
\multicolumn{9}{c}{\textsc{Token Pruning Rate = 88.9\%}} \\

SparseVLM
& 59.3
& \underline{64.2}
& 55.9
& 1690
& 83.3
& 75.7
& \textbf{58.8}
& 93.7 \\

PDrop
& 56.4
& 63.4
& 56.2
& 1663
& 77.6
& 73.5
& 54.4
& 90.4 \\

VisionZip
& 59.7
& 61.9
& 49.6
& 1643.7
& 83.2
& 72.8
& \underline{58.7}
& 90.9 \\

ApET
& \underline{61.0}
& 63.5
& \underline{56.6}
& \textbf{1783}
& \textbf{85.6}
& \underline{75.8}
& 54.4
& \underline{94.2} \\

\rowcolor{orange!8}
\proposed
& \textbf{61.9}
& \textbf{65.7}
& \textbf{57.4}
& \underline{1736}
& \underline{85.0}
& \textbf{77.1}
& \underline{58.7}
& \textbf{95.8} \\

\midrule

\rowcolor{gray!10}
\multicolumn{9}{c}{\textsc{Token Pruning Rate = 94.4\%}} \\

SparseVLM
& 51.2
& 52.1
& 48.6
& 1542
& 72.7
& 66.3
& 46.4
& 80.2 \\

PDrop
& 54.9
& \textbf{61.8}
& \textbf{54.9}
& 1513
& 72.3
& 70.2
& 52.7
& 86.4 \\

VisionZip
& 55.5
& 60.1
& 47.1
& 1628
& 74.8
& 71.4
& \textbf{56.2}
& 86.7 \\

ApET
& \underline{58.4}
& 60.8
& \underline{52.3}
& \underline{1680}
& \underline{82.6}
& \textbf{72.7}
& \underline{53.8}
& \underline{90.1} \\

\rowcolor{orange!8}
\proposed
& \textbf{58.9}
& \underline{61.7}
& 51.6
& \textbf{1701}
& \textbf{83.2}
& \underline{72.3}
& 53.1
& \textbf{90.2} \\

\bottomrule
\end{tabular}
\end{adjustbox}
\end{table*}

\subsubsection{Results on InternVL3-8B}
\label{sec:internvl-results}

 We evaluate \proposed on InternVL3-8B, whose high-resolution images
are represented by multiple visual tiles. This setting provides an additional
test of whether the sequential policy remains effective under a visual-token
organization different from both LLaVA and Qwen2.5-VL.

As shown in Table~\ref{tab:internvl-main}, \proposed achieves 95.4\% and
90.8\% average retention at 80.0\% and 90.0\% pruning, improving over ERA by
0.7 and 0.9 percentage points, respectively. The gains are particularly
consistent on TextVQA and GQA, where \proposed ranks first at both operating
points. Although it does not dominate every benchmark, the overall advantage
is maintained under both compression levels. Together with the results above,
this confirms that the learned selection policy transfers across substantially
different visual-token organizations.

\begin{table*}[!t]
\centering
\caption{Image-domain comparison with InternVL3-8B. ``Avg.'' is the mean
ratio of each benchmark score to the full-prefix upper bound. Best and
second-best pruned results at each pruning rate are shown in bold and
underlined, respectively.}
\label{tab:internvl-main}

\begin{adjustbox}{max width=\linewidth}
\begin{tabular}{lccccccc}
\toprule
Method &
TextVQA &
MME &
POPE &
GQA &
MMB &
MMBCN &
Avg. (\%) \\
\midrule

\rowcolor{gray!10}
\multicolumn{8}{c}{\textsc{Upper Bound: 1,280 Tokens (100\%)}} \\

Vanilla
& 81.5
& 2389.7
& 90.3
& 52.0
& 85.9
& 85.4
& 100.0 \\

\midrule

\rowcolor{gray!10}
\multicolumn{8}{c}{\textsc{Token Pruning Rate = 80.0\% }} \\

DivPrune
& 67.1
& 2188.7
& \textbf{90.1}
& 49.0
& 82.0
& 80.8
& 93.0 \\

VisionZip
& 67.2
& 2192.0
& 88.6
& 47.8
& \underline{82.9}
& 81.5
& 92.7 \\

DART
& 62.0
& \textbf{2280.4}
& 88.0
& \underline{50.9}
& \textbf{83.2}
& \textbf{83.2}
& 93.5 \\

ERA
& \underline{71.2}
& 2251.7
& \underline{89.8}
& 48.7
& \underline{82.9}
& \underline{83.0}
& \underline{94.7} \\

\rowcolor{orange!8}
\proposed
& \textbf{73.0}
& \underline{2264.5}
& \textbf{90.1}
& \textbf{51.2}
& \underline{82.9}
& 79.6
& \textbf{95.4} \\

\midrule

\rowcolor{gray!10}
\multicolumn{8}{c}{\textsc{Token Pruning Rate = 90.0\% }} \\

DivPrune
& 54.7
& 2043.8
& 87.6
& 47.4
& 78.6
& 78.0
& 87.3 \\

VisionZip
& 48.3
& 1987.8
& 85.2
& 44.6
& 77.7
& 76.5
& 83.8 \\

DART
& 51.3
& \textbf{2180.4}
& 85.1
& \underline{48.9}
& \textbf{80.7}
& \textbf{80.9}
& 88.5 \\

ERA
& \underline{60.8}
& 2119.2
& \textbf{88.7}
& 47.7
& \underline{79.6}
& \underline{79.8}
& \underline{89.9} \\

\rowcolor{orange!8}
\proposed
& \textbf{62.6}
& \underline{2144.7}
& \underline{88.3}
& \textbf{50.5}
& 79.5
& 77.5
& \textbf{90.8} \\

\bottomrule
\end{tabular}
\end{adjustbox}
\end{table*}

\begin{figure*}[!t]
    \centering
    \includegraphics[width=\textwidth]{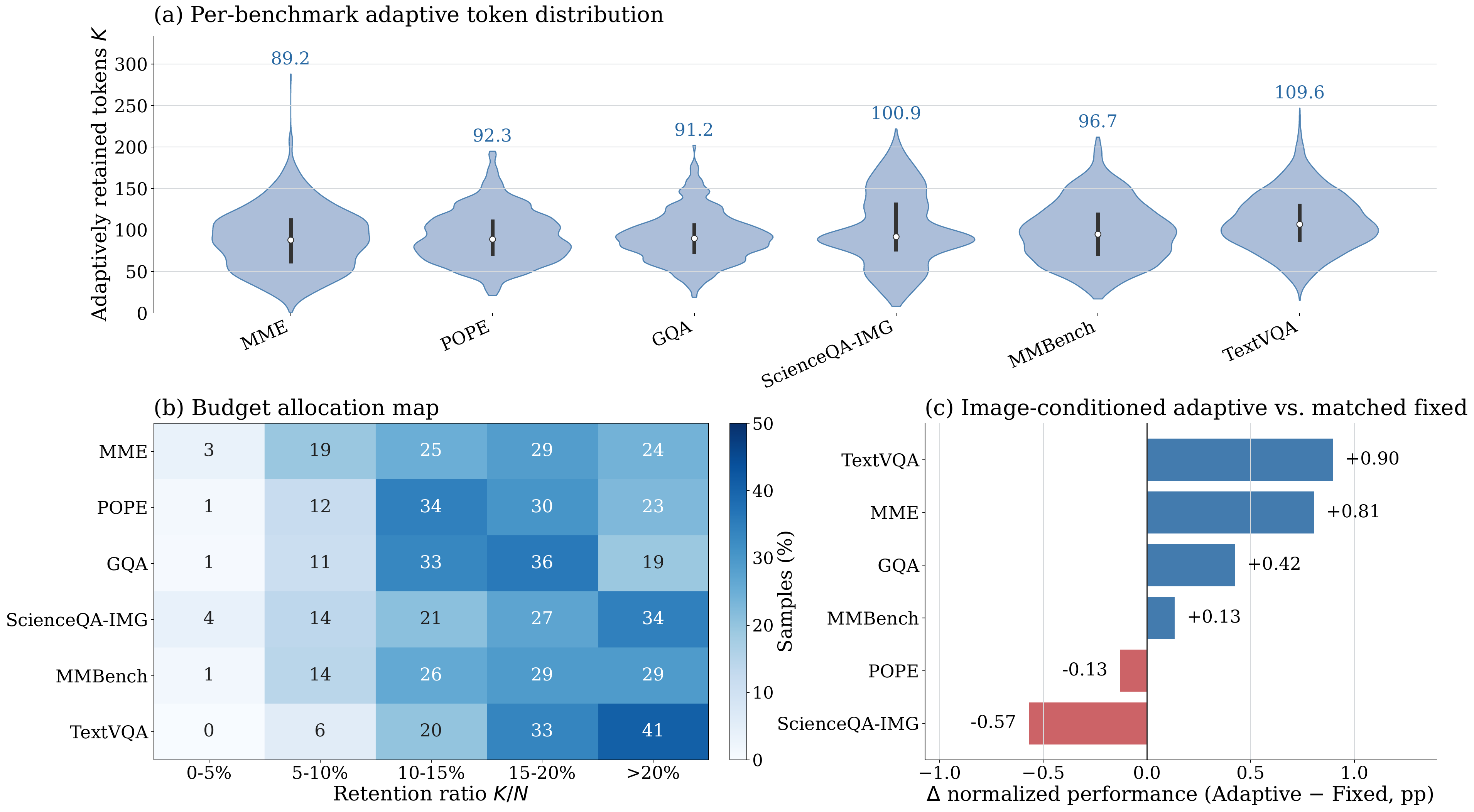}
    \caption{Adaptive token allocation across multimodal benchmarks.
    (a) Distribution of the adaptively retained token count $K$ on each
    benchmark, where the value above each distribution denotes the mean
    retained count. (b) Percentage of samples assigned to different retention
    intervals. (c) Difference in normalized performance between adaptive
    selection and fixed-budget selection under matched mean retained counts;
    positive values indicate an advantage for adaptive selection}
    \label{fig:adaptive}
\end{figure*}

\subsection{Adaptive Selection Analysis}
\label{sec:adaptive}

We analyze the adaptive behavior of the learned STOP action and the resulting
selection trajectories. The analysis uses LLaVA-1.5-7B on MME, POPE, GQA,
ScienceQA-IMG, MMBench, and TextVQA. Figure~\ref{fig:adaptive} reports the
retained-token distributions and the matched-budget comparison with fixed
selection.

\paragraph{Adaptive Budget Allocation.}
Figure~\ref{fig:adaptive} shows clear differences in retained budgets across
benchmarks. TextVQA has the largest mean retained count at 109.6, followed by
ScienceQA-IMG at 100.9. In contrast, MME, GQA, and POPE retain around 90
tokens on average. Variation is also observed within each benchmark. More than
20\% of the visual tokens are retained for 41\% of TextVQA samples, whereas
most GQA and POPE samples fall in the 10--20\% interval.

We next compare adaptive and fixed selection at the same mean retained count.
Adaptive selection improves TextVQA and MME by 0.90 and 0.81 percentage
points, respectively. The gains on GQA and MMBench are smaller, while POPE
and ScienceQA-IMG show slight decreases. The benefit is therefore not uniform
across tasks. The STOP action allows the retained length to vary from one
sample to another.

\begin{figure*}[!t]
    \centering
    \includegraphics[width=0.96\textwidth]{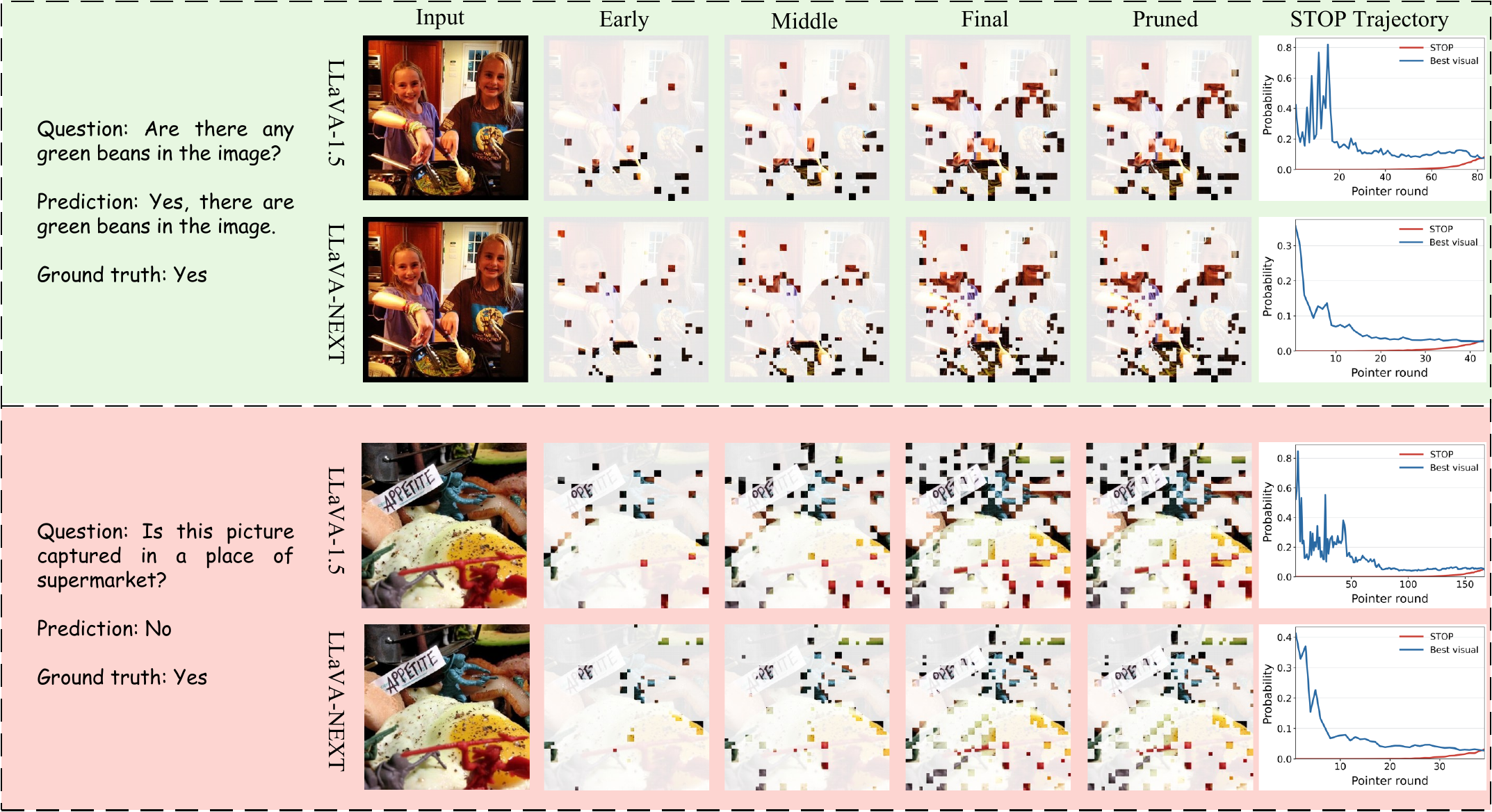}
    \caption{Sequential selection trajectories on LLaVA-1.5-7B and
    LLaVA-NeXT-7B. Columns show the input, cumulative selections at early,
    middle, and final stages, the final retained subset, and the STOP
    trajectory. The upper example is predicted correctly, while the lower
    example is a failure case.}
    \label{fig:selection-dynamics}
\end{figure*}

\paragraph{Sequential Selection Dynamics.}
Figure~\ref{fig:selection-dynamics} visualizes representative selection
trajectories. The examples are taken from LLaVA-1.5-7B and LLaVA-NeXT-7B.
The early, middle, and final columns show the cumulative selected regions.
The rightmost curves plot the STOP probability and the largest probability
among the remaining visual candidates.

The selected regions accumulate over successive steps. Each new selection is
conditioned on the subset retained in previous steps. The STOP trajectory
records when the policy terminates. In the failure case, the policy terminates
before all relevant local evidence is retained. Premature stopping therefore
remains a failure mode under aggressive compression.

\subsection{Efficiency and Scaling Analysis}
\label{sec:efficiency}

We evaluate the overhead of the recurrent pointer selector and its impact on
the efficiency gains from visual-prefix reduction.

\paragraph{Overall Efficiency.}
Table~\ref{tab:efficiency-comparison} compares FLOPs, latency, KV-cache
usage, GPU memory, and normalized performance retention. On LLaVA-1.5-7B,
\proposed reduces prefill latency from 59.95 ms to 40.05 ms at a mean
retained count of 64. This corresponds to a $1.50\times$ prefill speed-up
while retaining 94.6\% of the full-prefix performance. End-to-end latency
also decreases from 138.37 ms to 98.97 ms. At 128 retained tokens,
\proposed retains 96.7\% performance with a prefill latency of 50.08 ms.

The longer LLaVA-NeXT prefix yields larger prefill savings. At 160 and
320 retained tokens, \proposed achieves $3.61\times$ and $3.05\times$
prefill speed-ups over the full-prefix model, respectively. VisionZip remains
faster in absolute latency at the matched budgets. At 160 tokens, however,
\proposed uses fewer FLOPs, a smaller KV cache, and less GPU memory while
retaining 88.2\% normalized performance. The additional selector cost
therefore does not offset the measured prefill savings.

\begin{table*}[!t]
\centering
\caption{Efficiency and normalized performance retention on LLaVA-1.5-7B and
LLaVA-NeXT-7B under different retained-token budgets. Within each retained-token
setting, the best and second-best pruned results for each metric are shown in
bold and underlined, respectively. Lower is better for FLOPs, latency, KV cache,
and GPU memory, whereas higher is better for average performance retention.}
\label{tab:efficiency-comparison}

\resizebox{\textwidth}{!}{%
\begin{tabular}{lcccccccc}
\toprule
Method &
\begin{tabular}[c]{@{}c@{}}Token\\(\#)\end{tabular} &
\begin{tabular}[c]{@{}c@{}}FLOPs\\(TFLOPs)\end{tabular} &
\begin{tabular}[c]{@{}c@{}}E2E Latency\\(ms)\end{tabular} &
\begin{tabular}[c]{@{}c@{}}Prefill\\(ms)\end{tabular} &
\begin{tabular}[c]{@{}c@{}}Decode / Token\\(ms)\end{tabular} &
\begin{tabular}[c]{@{}c@{}}KV Cache\\(GB)\end{tabular} &
\begin{tabular}[c]{@{}c@{}}GPU Memory\\(GB)\end{tabular} &
\begin{tabular}[c]{@{}c@{}}Avg.\\(\%)\end{tabular} \\
\midrule


\rowcolor{gray!10}
\multicolumn{9}{c}{\textsc{Baseline: Full Model}} \\

LLaVA-1.5-7B
& 576
& 8.772
& 138.37
& 59.95
& 19.59
& 0.3350
& 15.296
& 100.0 \\

\rowcolor{gray!10}
\multicolumn{9}{c}{\textsc{Lightweight Methods (Retain 64 Tokens)}} \\

PDrop
& 64
& 4.690
& 135.84
& 49.37
& 21.46
& 0.1705
& 14.764
& 84.7 \\

VisionZip
& 64
& \textbf{1.984}
& 140.98
& 50.13
& 22.65
& \textbf{0.0666}
& \textbf{14.588}
& \underline{92.0} \\

SparseVLM
& 64
& 2.253
& 124.29
& 41.29
& 20.67
& 0.0762
& 18.919
& 85.6 \\

\rowcolor{orange!8}
\proposed
& 64
& 2.026
& \textbf{98.97}
& \textbf{40.05}
& \textbf{14.73}
& \textbf{0.0666}
& 14.707
& \textbf{94.6} \\

\rowcolor{gray!10}
\multicolumn{9}{c}{\textsc{Lightweight Methods (Retain 128 Tokens)}} \\

PDrop
& 128
& 5.415
& 100.09
& 43.20
& 14.23
& 0.1986
& 14.870
& 95.5 \\

VisionZip
& 128
& \textbf{2.832}
& \textbf{84.57}
& \textbf{29.51}
& \textbf{13.77}
& \textbf{0.1001}
& \textbf{14.621}
& \underline{96.4} \\

SparseVLM
& 128
& 3.089
& 129.40
& 47.72
& 20.39
& 0.1093
& 18.930
& 92.6 \\

\rowcolor{orange!8}
\proposed
& 128
& 2.875
& 107.23
& 50.08
& 14.31
& \textbf{0.1001}
& 14.827
& \textbf{96.7} \\

\midrule


\rowcolor{gray!10}
\multicolumn{9}{c}{\textsc{Baseline: Full Model}} \\

LLaVA-NeXT-7B
& 2880
& 48.473
& 417.89
& 354.51
& 15.86
& 0.3878
& 17.952
& 100.0 \\

\rowcolor{gray!10}
\multicolumn{9}{c}{\textsc{Lightweight Methods (Retain 160 Tokens)}} \\

PDrop
& 160
& 19.737
& 216.15
& 134.83
& 20.16
& 0.1550
& 17.658
& 86.4 \\

VisionZip
& 160
& 5.049
& \textbf{129.38}
& \textbf{50.75}
& 19.79
& 0.0304
& 16.658
& \textbf{88.2} \\

SparseVLM
& 160
& 11.347
& 162.99
& 82.68
& 20.19
& 0.0840
& 17.598
& 80.2 \\

\rowcolor{orange!8}
\proposed
& 160
& \textbf{4.783}
& 171.93
& 98.17
& \textbf{16.60}
& \textbf{0.0290}
& \textbf{15.901}
& \textbf{88.2} \\

\rowcolor{gray!10}
\multicolumn{9}{c}{\textsc{Lightweight Methods (Retain 320 Tokens)}} \\

PDrop
& 320
& 22.345
& 230.20
& 150.14
& 19.97
& 0.1777
& 17.658
& 90.4 \\

VisionZip
& 320
& 7.586
& \textbf{141.48}
& \textbf{66.65}
& 18.67
& 0.0523
& 16.658
& \textbf{93.7} \\

SparseVLM
& 320
& 14.085
& 175.72
& 94.51
& 20.39
& 0.1083
& 17.598
& \textbf{93.7} \\

\rowcolor{orange!8}
\proposed
& 320
& \textbf{7.111}
& 186.33
& 116.30
& \textbf{16.49}
& \textbf{0.0503}
& \textbf{15.901}
& 92.1 \\

\bottomrule
\end{tabular}%
}
\end{table*}

\paragraph{Scaling with Visual-Prefix Length.}
We further profile the selector and prefill cost as the visual-prefix length
increases from 320 to 2240 tokens under 10\% retention.
Table~\ref{tab:scaling} shows that selector latency increases with sequence
length, but full-prefix prefill grows more rapidly.

On LLaVA-1.5-7B, the measured prefill speed-up increases from
$0.98\times$ at 320 input tokens to $4.10\times$ at 2240 tokens.
On LLaVA-NeXT-7B, it increases from $1.24\times$ to $3.88\times$.
The selector share reaches 67.80\% and 65.34\% at 2240 tokens,
respectively. The selector occupies a larger fraction of the compressed
prefill path as the sequence grows. However, full-prefix prefill grows
faster, leading to larger net speed-ups.

\begin{table*}[!t]
\centering
\caption{Scaling with input token count at 10\% retention on a single
NVIDIA RTX PRO 6000 Blackwell GPU. Latencies are in milliseconds}
\label{tab:scaling}

\resizebox{\textwidth}{!}{%
\begin{tabular}{llrrrrrrr}
\toprule
Backbone & Input tokens & 320 & 640 & 960 & 1280 & 1600 & 1920 & 2240 \\
\midrule

\multirow{6}{*}{LLaVA-1.5-7B}
& Retained tokens
& 32 & 64 & 96 & 128 & 160 & 192 & 224 \\

& Full prefill
& 29.34 & 53.24 & 70.86 & 116.89 & 142.93 & 189.83 & 236.55 \\

& \proposed\ prefill
& 30.00 & 37.14 & 37.75 & 40.91 & 46.13 & 52.28 & 57.73 \\

& Speed-up
& $0.98\times$ & $1.43\times$ & $1.88\times$ & $2.86\times$
& $3.10\times$ & $3.63\times$ & $4.10\times$ \\

& Selector latency
& 9.65 & 16.15 & 20.45 & 24.25 & 28.85 & 34.41 & 39.14 \\

& Selector share
& 32.16\% & 43.49\% & 54.17\% & 59.28\%
& 62.54\% & 65.81\% & 67.80\% \\

\midrule

\multirow{6}{*}{LLaVA-NeXT-7B}
& Retained tokens
& 32 & 64 & 96 & 128 & 160 & 192 & 224 \\

& Full prefill
& 29.57 & 47.11 & 75.46 & 119.36 & 150.81 & 187.45 & 239.61 \\

& \proposed\ prefill
& 23.91 & 29.99 & 37.96 & 43.40 & 49.84 & 56.32 & 61.77 \\

& Speed-up
& $1.24\times$ & $1.57\times$ & $1.99\times$ & $2.75\times$
& $3.03\times$ & $3.33\times$ & $3.88\times$ \\

& Selector latency
& 7.25 & 12.61 & 19.08 & 23.78 & 29.48 & 35.46 & 40.36 \\

& Selector share
& 30.30\% & 42.05\% & 50.27\% & 54.80\%
& 59.15\% & 62.97\% & 65.34\% \\

\bottomrule
\end{tabular}%
}
\end{table*}

\subsection{Component Effectiveness Verification}
\label{sec:ablation}

We analyze the main components of \proposed on LLaVA-1.5-7B.
Table~\ref{tab:component-ablation} reports ablations of the cross-modal
encoder and trajectory aggregation rule on the eight benchmarks in
Table~\ref{tab:llava-main}. Train--inference alignment is evaluated on GQA,
MMB, POPE, and MME.

\begin{table}[!t]
\centering
\caption{Component ablations on LLaVA-1.5-7B. Each row reports average
performance retention (\%) over the eight benchmarks in
Table~\ref{tab:llava-main}. Rows marked with $\dagger$ indicate the default
configuration}
\label{tab:component-ablation}

\begin{tabular}{lc}
\toprule
Variant & Avg. Ret. (\%) \\
\midrule

\rowcolor{gray!10}
\multicolumn{2}{l}{\textit{(a) Encoder self-attention direction}} \\

Causal (unidirectional) & 92.4 \\
Bidirectional$^\dagger$ & 94.6 \\

\midrule

\rowcolor{gray!10}
\multicolumn{2}{l}{
\textit{(b) Scorer aggregation (see Eq.~\eqref{eq:aggregate})}
} \\

Last-step pointer only & 90.3 \\
Uniform sum ($A_t\equiv1,\ \beta=1$) & 83.9 \\
Alive-weighted only ($\beta=1$) & 89.5 \\
$A_t\cdot p_t(i)\cdot\beta^t$ $^\dagger$ & 94.7 \\

\bottomrule
\end{tabular}
\end{table}

\paragraph{Encoder Self-Attention Direction.}
The cross-modal encoder constructs the candidate memory before sequential
decoding begins. At this stage, the full visual and textual context is already
available. A causal mask therefore restricts interactions according to token
order. Bidirectional self-attention achieves 94.6\% average retention,
compared with 92.4\% for causal attention. The 2.2-point difference favors
bidirectional contextualization before sequential selection.

\paragraph{Trajectory Aggregation.}
The differentiable score in Eq.~\eqref{eq:aggregate} combines the alive
probability $A_t$, pointer probability $p_t(i)$, and geometric factor
$\beta^t$. Table~\ref{tab:component-ablation} isolates their effects.
Using only the final pointer distribution retains 90.3\%, while an unweighted
sum over all steps drops to 83.9\%. Preserving the alive probability but
removing the geometric decay reaches 89.5\%. The full aggregation rule
achieves the highest retention of 94.7\%. Both trajectory weighting terms
therefore contribute to the training signal.

\paragraph{Train--Inference Alignment.}
Training uses the continuous noise-gated representation, whereas inference
physically gathers the hard-selected visual tokens. We therefore evaluate
both paths with the same trained selector. As reported in
Table~\ref{tab:train-infer-diagnostic}, hard pruning retains 97.3\%, 95.8\%,
and 93.7\% at approximately 192, 128, and 64 tokens, respectively.
Noise-gated inference retains 97.0\%, 94.9\%, and 92.5\% at the same
operating points. The corresponding gaps range from 0.3 to 1.2 percentage
points.

\begin{table}[!t]
\centering
\caption{Train--inference diagnostic on LLaVA-1.5-7B. Values are average
performance retention (\%) over GQA, MMB, POPE, and MME}
\label{tab:train-infer-diagnostic}
\setlength{\tabcolsep}{3.5pt}
\begin{tabular}{lccc}
\toprule
Variant & $K\approx192$ & $K\approx128$ & $K\approx64$ \\
\midrule
Hard pruning & 97.3 & 95.8 & 93.7 \\
Noise-gated inference & 97.0 & 94.9 & 92.5 \\
Gap & 0.3 & 0.9 & 1.2 \\
\bottomrule
\end{tabular}
\end{table}

The gap increases from 0.3 to 1.2 points as the retained budget decreases.
Hard pruning is consistently stronger than the noise-gated diagnostic path.
The diagnostic results therefore do not overstate the reported hard-pruning
performance.

\subsection{Sensitivity Analysis}
\label{sec:sensitivity}

\begin{figure*}[!t]
\centering
\includegraphics[width=\textwidth]{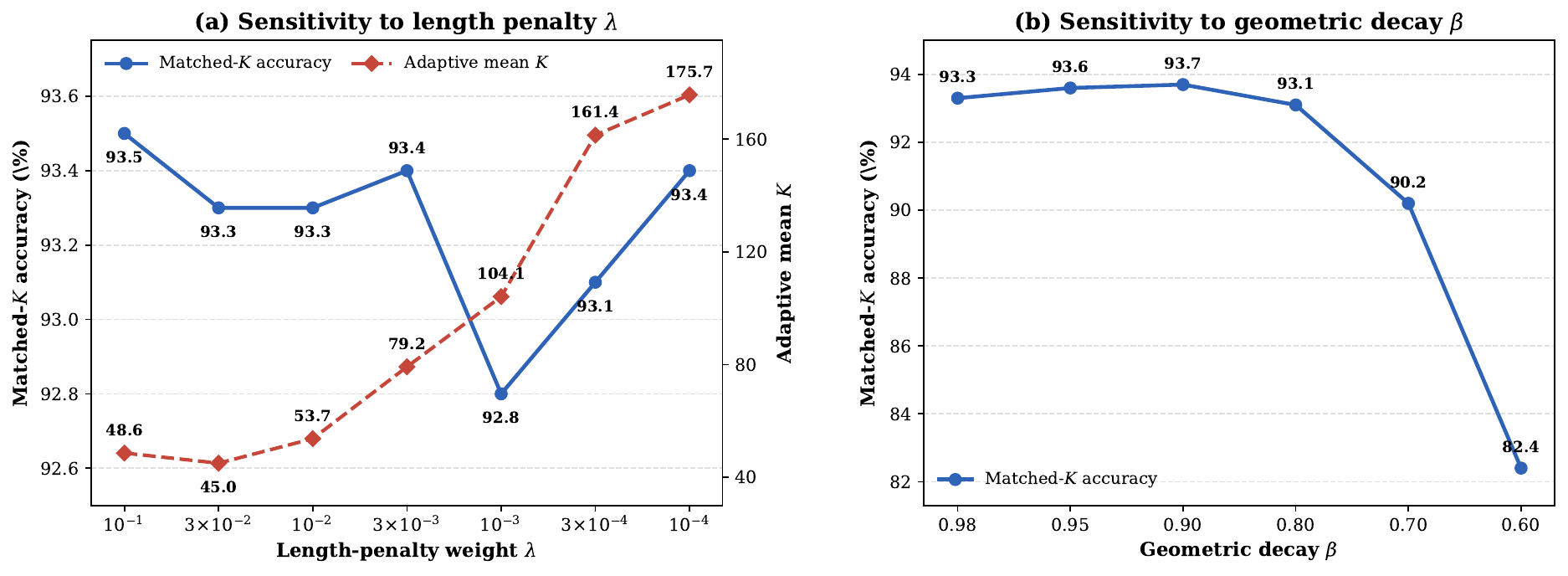}
\caption{Hyperparameter sensitivity on LLaVA-1.5-7B.
(a) Adaptive mean retained count $K$ and matched-$K$ retention under
different length-penalty weights $\lambda$.
(b) Matched-$K$ retention under different geometric decay factors $\beta$.
Retention is averaged over GQA, MMB, POPE, and MME at matched
$K\approx64$.}
\label{fig:hyperparameter-sensitivity}
\end{figure*}

We analyze the sensitivity to the length-penalty weight $\lambda$ and
geometric decay $\beta$. Figure~\ref{fig:hyperparameter-sensitivity} reports
the adaptive mean retained count and matched-budget performance.

Varying $\lambda$ primarily changes the retained token count. Under stronger
length pressure, the adaptive mean count falls to 45.0--53.7 tokens, whereas
weaker penalties allow the mean to exceed 160 tokens. At matched
$K\approx64$, normalized retention remains within 92.8--93.5\%. This suggests
that $\lambda$ mainly controls when the policy stops, while having a smaller
effect on token ranking at a matched budget.

For $\beta$, performance is stable over a broad middle range. Matched-budget
retention remains between 93.1\% and 93.7\% for
$\beta\in[0.8,0.98]$. Reducing $\beta$ to 0.70 and 0.60 lowers retention to
90.2\% and 82.4\%, respectively. Small $\beta$ values place more weight on
the earliest pointer steps and reduce the contribution of later selections.
Performance therefore remains stable for moderate $\beta$ values, while
overly strong decay causes a clear drop.

\section{Discussion and Limitations}
\label{sec:discussion}

\subsection{Sequential Selection versus One-Shot Ranking}

The main difference between \proposed and one-shot pruning lies in how the
retained subset is constructed. One-shot methods score all visual tokens
before selection and retain the highest-ranked candidates. Their scores are
not updated after part of the visual evidence has been selected. In contrast,
\proposed recomputes the pointer distribution after each selection step. The
remaining candidates are therefore evaluated under the updated selection
history.

This distinction is useful when the required evidence is distributed across
multiple regions. A selected token may reduce the value of redundant
candidates or increase the relevance of complementary ones. The STOP action
further allows the process to terminate without a predefined per-image token
count. However, sequential updates are not expected to benefit every input
equally. For visually simple or redundant inputs, a strong one-shot ranking
may already provide an adequate subset. The advantage of sequential selection
therefore depends on the structure of the visual evidence.

\subsection{Adaptive STOP and Task Difficulty}

The retained count $K$ varies across inputs, but it should not be interpreted
as a universal measure of semantic difficulty. It depends on the backbone,
prompt, trained selector, and length penalty. Figure~\ref{fig:adaptive} shows
clear differences in the retained-token distributions across benchmarks.
TextVQA and ScienceQA-IMG receive larger mean budgets than MME, GQA, and POPE.
This pattern is consistent with the greater demand for fine-grained or
distributed evidence in some inputs.

The global operating point and the per-sample budget are also different.
The length penalty determines the overall retention regime during training,
whereas the STOP action produces different values of $K$ within that regime.
Figure~\ref{fig:hyperparameter-sensitivity} shows that changing the length
penalty substantially shifts the mean retained count, while matched-budget
performance changes much less. Thus, inference does not require a fixed token
count for every input. However, different global efficiency--accuracy regimes
still rely on separately trained selectors. Supporting multiple operating
points with a single checkpoint remains an open direction.

\subsection{Scalability and Practical Overhead}

Sequential selection adds selector computation before language-model prefill.
Its practical value therefore depends on the amount of prefill computation
saved after pruning. Table~\ref{tab:scaling} shows that selector latency
becomes a larger fraction of the compressed prefill path as the visual
sequence grows. Full-prefix prefill, however, grows more rapidly. The measured
net speed-up consequently increases over the tested range.

This trend is particularly relevant to long visual prefixes. The
LLaVA-NeXT results show larger prefill gains than the shorter-prefix setting,
suggesting greater practical value for high-resolution representations.
Similar benefits may extend to dynamic-resolution and multi-tile models, but
their scaling behavior is not profiled here.

The same experiment also shows the boundary of this trade-off. At short
sequence lengths, selector overhead can offset much of the prefill saving. The
320-token LLaVA-1.5 setting in Table~\ref{tab:scaling} provides such an
example. Grouped execution reduces this overhead for longer prefixes by
selecting multiple tokens per pointer round. It also reduces selection
granularity because the history is updated between groups rather than after
every token. The group size therefore trades fine-grained conditioning for
lower selector latency.

\subsection{Limitations and Future Work}

Several limitations remain. First, fine-grained local evidence is still
difficult to preserve under aggressive compression. This is most visible for
text-rich inputs and in the failure case of
Fig.~\ref{fig:selection-dynamics}. If STOP is triggered too early, relevant
local evidence may never enter the retained subset. More reliable stopping
criteria could reduce this failure mode.

Second, training and inference use different representations. Training keeps
a full-length noise-gated sequence, whereas inference physically removes
unselected tokens. Table~\ref{tab:train-infer-diagnostic} shows a small gap
between the two paths, and the gap increases as the retained budget becomes
smaller. A more direct optimization of hard selection may further reduce this
difference.

Finally, the current evaluation focuses on image benchmarks and
7B--8B-scale open MLLMs. The tested backbones cover fixed-resolution, AnyRes,
dynamic-resolution, and multi-tile visual representations, but do not include
video inputs or substantially larger language models. Video introduces both
spatial and temporal redundancy and may require a hierarchical selection
strategy. Cross-backbone selector transfer and multi-budget inference also
remain open problems.

Despite these limitations, these properties also suggest promising directions for future work in high-resolution visual domains. In particular, medical imaging and related settings are characterized by substantial redundancy in visual tokens~\cite{yang2025one,young2026fewer,young2026scalar}. This phenomenon has been widely observed in medical image analysis~\cite{yang2024segmentation,yang2023geometry,chen2026tc,wu2026multimodal,xu2026unified}, multimodal medical foundation models~\cite{xu2023learning,xu2024medvilam,xu2024foundation,feng2026efficient}, and visual prune tasks~\cite{he2026diffprune,he2026beyond,he2026autoselect,he2026stepwise,chen2026learnable,chen2026pathselect,gao2026decoupling,gao2026zerosense}, suggesting that StepPrune may be especially effective in these scenarios.

\section{Conclusion}
\label{sec:conclusion}

\proposed formulates visual-token pruning as history-conditioned sequential
selection. A pointer decoder progressively constructs the retained subset,
while a learned STOP action determines its input-adaptive size. During
training, a noise-gated surrogate provides gradients for the discrete
selection process. During inference, unselected visual tokens are removed
before language-model prefill.

Across the evaluated pruning rates, \proposed achieves the best average
normalized performance retention on LLaVA-1.5, Qwen2.5-VL, and InternVL3,
while remaining competitive on LLaVA-NeXT. On LLaVA-1.5, it retains 94.6\%
of the full-prefix normalized performance at 88.9\% pruning. With a mean
retained count of 64, prefill latency decreases from 59.95 ms to 40.05 ms,
corresponding to a $1.50\times$ speed-up. These results demonstrate the
potential of sequential and adaptive visual-token selection for reducing
MLLM inference cost.

\section*{Acknowledgments}

This research was partially supported by the Open Project Program of State Key Laboratory of Virtual Reality Technology and Systems, Beihang University (No.VRLAB2026B06) and the Shenzhen Medical Research Fund (Grant No. D260403015).

\section*{Contribution Statement}
Hansen Zhang and Landi He contributed equally to this work. Lijian Xu is the corresponding author.

\section*{Statements and Declarations}

\subsection*{Data Availability}

The datasets used in this study are publicly available and can be accessed through their respective official repositories or project pages.

\subsection*{Competing Interests}

The authors have no competing interests to declare that are relevant to the
content of this article.

\FloatBarrier

\bibliographystyle{named}
\bibliography{references}

\end{document}